\documentclass[conference]{IEEEtran}
\IEEEoverridecommandlockouts
\usepackage{cite}
\usepackage{amsmath,amssymb,amsfonts}
\usepackage{algorithmic}
\usepackage{graphicx}
\usepackage{textcomp}
\usepackage{xcolor}
\def\BibTeX{{\rm B\kern-.05em{\sc i\kern-.025em b}\kern-.08em
    T\kern-.1667em\lower.7ex\hbox{E}\kern-.125emX}}

\usepackage{adjustbox}
\usepackage{booktabs}
\usepackage{hyperref}
\usepackage{multirow}
\usepackage[disable]{todonotes}
\usepackage{subcaption}
\begin{document}

\newcommand{\convtwo}{\textsc{Conv}}
\newcommand{\seqnca}{\textsc{SeqNCA}}
\newcommand{\conv}{\textsc{Conv}}

\newcommand{\binary}{\textsc{binary}}
\newcommand{\maze}{\textsc{maze}}
\newcommand{\dungeon}{\textsc{dungeon}}

\title{Scaling, Control and Generalization in Reinforcement Learning Level Generators
}



\author{\IEEEauthorblockN{Sam Earle}
\IEEEauthorblockA{\textit{New York University} \\
\textit{Game Innovation Lab}\\
Brooklyn, NY \\
sam.earle@nyu.edu}
\and
\IEEEauthorblockN{Zehua Jiang}
\IEEEauthorblockA{\textit{New York University} \\
\textit{Game Innovation Lab}\\
Brooklyn, NY \\
zehua.jiang@nyu.edu}
\and
\IEEEauthorblockN{Julian Togelius}
\IEEEauthorblockA{\textit{New York University} \\
\textit{Game Innovation Lab}\\
Brooklyn, NY \\
julian@togelius.com}
}

\maketitle
\IEEEoverridecommandlockouts
\IEEEpubid{\makebox[\columnwidth]{ 979-8-3503-5067-8/24/\$31.00~\copyright2024 IEEE \hfill} 
\hspace{\columnsep}\makebox[\columnwidth]{ }}
\IEEEpubidadjcol

\begin{abstract}
Procedural Content Generation via Reinforcement Learning (PCGRL) has been introduced as a means by which controllable designer agents can be trained based only on a set of computable metrics acting as a proxy for the level's quality and key characteristics.
While PCGRL offers a unique set of affordances for game designers, it is constrained by the compute-intensive process of training RL agents, and has so far been limited to generating relatively small levels.
To address this issue of scale, we implement several PCGRL environments in Jax so that all aspects of learning and simulation happen in parallel on the GPU, resulting in faster environment simulation; removing the CPU-GPU transfer of information bottleneck during RL training; and ultimately resulting in significantly improved training speed.
We replicate several key results from prior works in this new framework, letting models train for much longer than previously studied, and evaluating their behavior after $1$ billion timesteps.
Aiming for greater control for human designers, we introduce randomized level sizes and frozen ``pinpoints'' of pivotal game tiles as further ways of countering overfitting.
To test the generalization ability of learned generators, we evaluate models on large, out-of-distribution map sizes, and find that partial observation sizes learn more robust design strategies.
\end{abstract}


\section{Introduction}
In procedural content generation via reinforcement learning (PCGRL), the process of iterative game design is frames as a markov decision process, and reinforcement learning (RL) agents are trained to generate game content. 
Instead of learning to play a game by taking actions, observing states, and getting rewards, these agents learn to generate (parts of) a game by taking actions, observing states, and getting rewards. The actions edit the content artifact, and the reward is based on the quality of the artifact that is being created.

The advantage of PCGRL is that you can use it to create not just game content, but game content \emph{generators}. Compared to search-based approaches, this means that almost all the compute is front-loaded; first you train the generator, then inference is fast and cheap. This makes it suitable for runtime use in games. Compared to supervised or self-supervised learning, PCGRL don't need any existing content to train on. This makes it suitable for use for games where any content has yet to be produced.

Despite these considerable potential advantages, PCGRL has only limited uptake since it was first proposed in~\cite{khalifa2020pcgrl}. This could be due to the difficulty of designing good reward functions, the tendency to overfit to single solutions, the long training time, and the problems with scaling to produce larger-size levels and other content. In this paper, we propose and evaluate several modifications to the basic PCGRL formulation aimed at rectifying some of these issues.

Two novel elements we propose are randomizing level size during training, and pinpointing locations of key elements. Both of these interventions function to limit overfitting by enforcing closed-loop policies, in other words, the agent must take its observations into account and cannot rely on rote-learning parts of levels.
These add new degrees of controllability in addition on top of conditioning on high-level features introduced in~\cite{earle2021learning}.

We also examine the effects of systematically changing the size of the agent's observation space. In the original PCGRL formulation, the observation window typically covers the whole level. From reinforcement learning experiments in various domains, we know that observation and structure can have large effects on overfitting and scalability. We hypothesize that the same is true for PCGRL, and that we can improve generalization and scalability by choosing adequate observation windows. This hypothesis is largely confirmed by our experiments.
We find that smaller observation windows always increase generalization to new level sizes.
On a task involving pinpoints and randomized map shapes during training, these more local models additionally perform comparably or better in-distribution.



To offset the high computational cost of training content-generating agents, we reimplement the standard PCGRL library in jax~\cite{jax2018github}, a framework that allows a high degree of parallelization using the GPU to simulate the environment, resulting for $15\times$ speedups during training, making it feasible to experiment with longer training times.


In sum, our contributions are as follows
\begin{itemize}
    \item We re-implement the PCGRL code-base in jax, making it practical to scale PCGRL to larger and more complex domains.
    \item We add new features---variable map shapes and varied (frozen) placement of pivotal ``pinpoint'' tiles---to make the task of level generation more complex, and to make the result generators more controllable.
    \item We conduct a thorough investigation of the effects of partial observations, finding that partial observations are more successful in generalizing to large map sizes unseen during training.
\end{itemize}

\begin{figure}
\begin{subfigure}{\linewidth} 
\begin{subfigure}{.49\linewidth} 
\includegraphics[width=\linewidth,trim={0 160 0 0},clip]{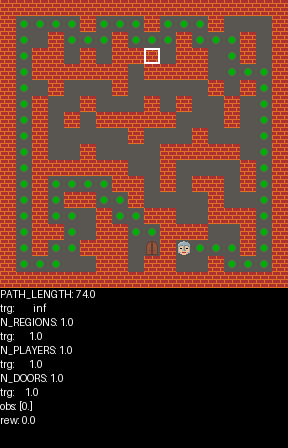}
\caption{$8\times 8$ observations}
\end{subfigure}
\begin{subfigure}{.49\linewidth} 
\includegraphics[width=\linewidth,trim={0 160 0 0},clip]{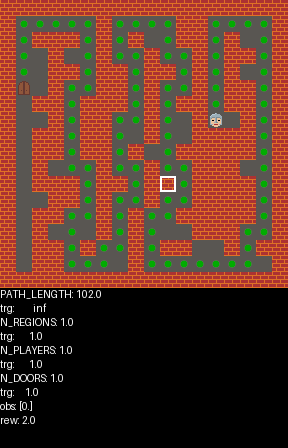}
\caption{$31\times 31$ (global) observations}
\end{subfigure}
\caption{Evaluation on in-distribution $16\times 16$ maps.}
\end{subfigure}
\begin{subfigure}{\linewidth} 
\vspace{10pt}
\begin{subfigure}{.49\linewidth} 
\includegraphics[width=\linewidth,trim={0 160 0 0},clip]{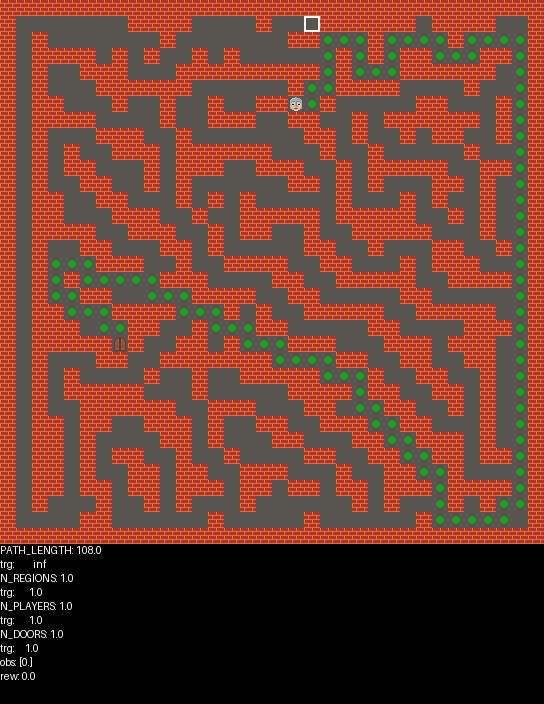}
\caption{$8\times 8$ observations}
\end{subfigure}
\begin{subfigure}{.49\linewidth} 
\includegraphics[width=\linewidth,trim={0 160 0 0},clip]{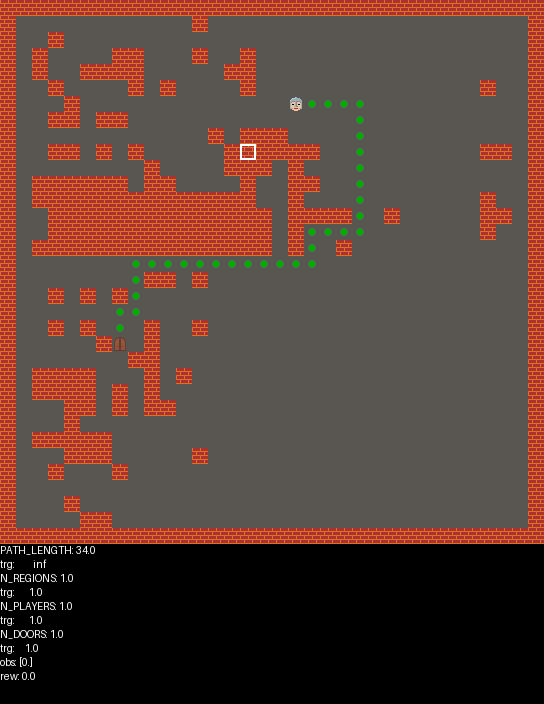}
\caption{$31\times 31$ (global) observations}
\end{subfigure}
\caption{Evaluation on out-of-distribution $32\times 32$ maps.}
\end{subfigure}
\caption{Evaluation in the \maze{} domain with pinpoints (randomly fixed player and door tiles). While models with large global observations are better on small $16\times 16$ in-distribution maps, models with smaller local observations learn scalable patterns that generalize better to larger $32\times 32$ maps.}
\label{diffshape_pinpoint_maze_emptystart_maps}
\end{figure}

\section{Background}

Video games featuring some form of content generation, often level generation, have existed since the early 1980s. Beneath Apple Manor, Rogue, and Elite were the early poster children of early game content generation. PCG has become a mainstay in modern games, with titles such as No Man's Sky, Hades, and the Civilization series that rely heavily on some form of runtime content generation.

Academic research in content generation dates back to the early 2000s~\cite{shaker2016procedural}. Much of the early research focused on search-based approaches~\cite{togelius2011search}. Search-based approaches are very versatile, but the computational demands at generation time can be high, making such approaches harder to use for runtime generation. Constraint satisfaction approaches~\cite{smith2011answer,smith2011tanagra} were also given considerable attention. While constraint satisfaction approaches can be very powerful, it imposes particular constraints on the shape of the content.

Supervised and self-supervised approaches to PCG started being explored seriously at the dawn of the deep learning era~\cite{summerville2018procedural,liu2021deep,guzdial2022procedural}. A variety of these machine learning methods have been applied to game content generation, including Generative Adversarial Networks~\cite{volz2018evolving}, LSTM networks~\cite{summerville2016super}, and Markov models~\cite{snodgrass2016learning}. However, these methods generally have large requirements on training data. This begs the question that has been called the fundamental tension of PCGML: \textit{if these methods only work well with enough existing content, why would you need to generate it?}~\cite{karth2019addressing}.

Reinforcement learning approaches to PCG are more recent, first proposed by~\cite{guzdial2018co} and~\cite{khalifa2020pcgrl}. There are significant advantages to PCGRL over existing approaches:
\begin{enumerate}
  \item No training data is necessary
  \item Generators are very fast during inference time
  \item The generator is iterative, allowing mixed-initiative solutions~\cite{delarosa2021mixed}
\end{enumerate}
With these advantages comes a unique set of considerations. In principle, the same kind of evaluations used as fitness functions for search-based PCG can be used as reward functions in PCGRL. However, due to longer training times, a computationally lightweight reward function is needed in settings with dense reward. There is also a tendency to mode collapse via overfitting, which can be counteracted via limiting the number of changes the model can make, or introducing conditional inputs~\cite{earle2021learning}. Because of these constraints, scaling PCGRL to larger-sized levels or artefacts has proven a challenge~\cite{jiang2022learning}.

One type of modification to the basic PCGRL formula explored here concerns the size and shape of the observation. This draws on earlier results showing that limiting the observation size and aligning it properly can greatly help with generalization~\cite{ye2020rotation}.
\todo[]{CITE Matt Fontaine scaling NCAs to larger levels}

\section{Methods}
\begin{figure}
\centering
\begin{subfigure}[t]{\linewidth}
\includegraphics[width=\linewidth]{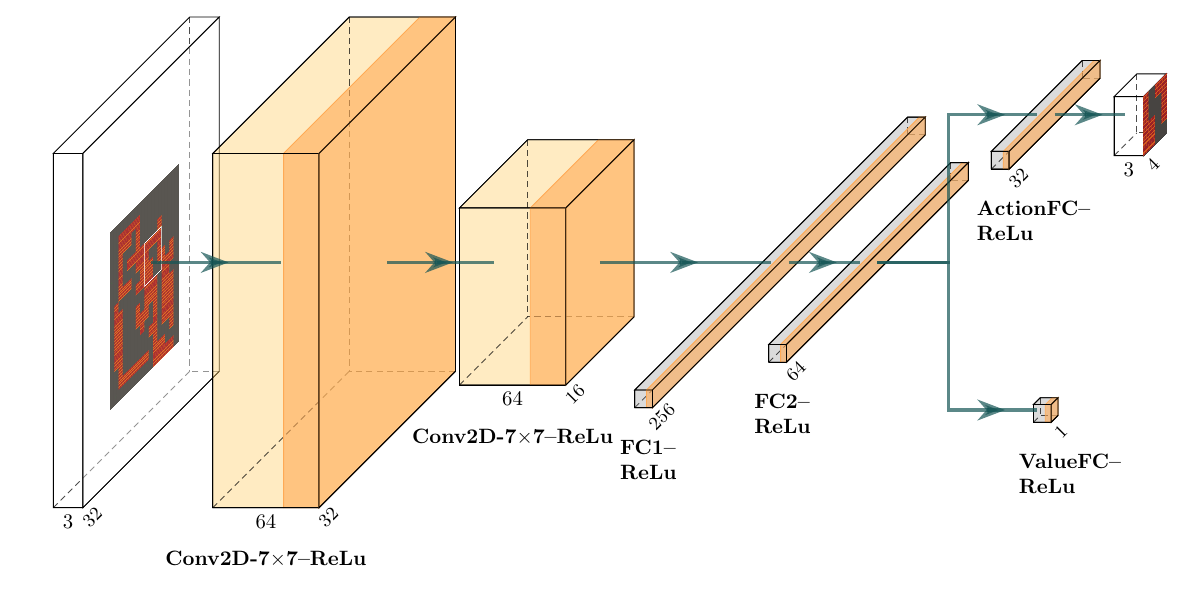}
\caption{\textbf{\convtwo{} model.} Both action and value branches process a single local or global 2D observation with convolutional and fully connected layers.}
\label{fig:default_arch}
\end{subfigure}
\caption{\textbf{Model architectures.}}
\end{figure}

\subsection{Training}



\begin{table}
    \centering
\adjustbox{max width=\linewidth}{%
\begin{tabular}{lllrrrrrrr}
& & n. envs & 1 & 10 & 50 & 100 & 200 & 400 & 600 \\ 
    \hline
  \multicolumn{2}{c}{CPU}  & binary & 189 & 420 & 520 & 524 & 525 & 529 & 522 \\ 
  &  & dungeon & 206 & 416 & 505 & 511 & 509 & 511 & 496 \\ 
\hline
\multicolumn{2}{c}{jax} & binary & 256 & 1,256 & 1,098 & 3,993 & 6,076 & 7,814 & 8,943 \\ 
& &    maze & 471 & 2,099 & 6,159 & 8,039 & 9,319 & 9,456 & 10,961  \\ 
&  &   dungeon & 292 & 1,463 & 4,042 & 5,375 & 6,851 & 7,502 & 8,798
\end{tabular}
}
    \caption{Environment steps per second on various domains in the CPU and jax implementations of~\cite{earle2021learning} during training, on 10 CPUs or an RTX-8000 respectively, under varying numbers of parallelized environments.}
    \label{tab:jax_train_fps}
\end{table}

\begin{table}
    \centering
\adjustbox{max width=\linewidth}{%
\begin{tabular}{lrrrrrrr}
n. envs &  50 & 100 & 200 & 400 & 600 \\
\midrule
binary & 6846.73 & 11979.56 & 21478.48 & 37072.85 & 46498.75 \\
maze &  6867.30 & 12325.76 & 21902.67 & 37745.91 & 47323.79 \\
dungeon &  6923.89 & 12349.95 & 21882.25 & 37770.10 & 46839.11 \\
\end{tabular}
}

    \caption{Environment steps per second on various domains in the jax implementation of PCGRL while taking random actions, on an RTX-8000, under varying numbers of parallelized environments. Frames per-second exceed 45k, despite the relative complexity of the pathfinding operations required to compute reward in these domains.}
    \label{tab:jax_fps}
\end{table}

\begin{figure}[t]
\begin{subfigure}{\linewidth}
    \includegraphics[width=\linewidth]{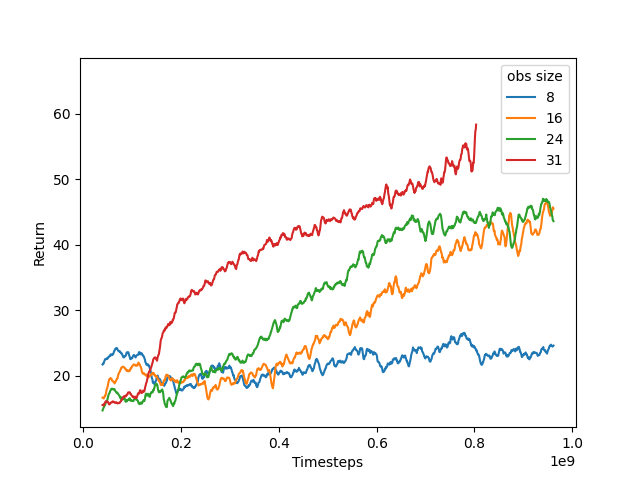}
    \caption{Fixed map shapes during training.}
    \label{fig:diffshape_pinpoint_maze_emptystart_metric_curves_mean_fixedShape}
\end{subfigure}
\begin{subfigure}{\linewidth}
    \includegraphics[width=\linewidth]{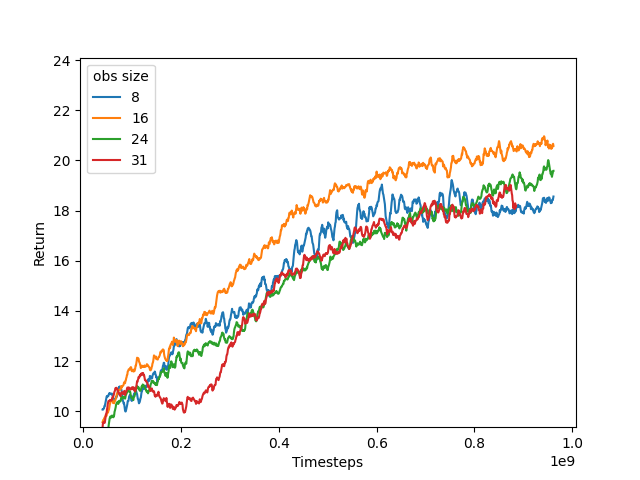}
    \caption{Randomized per-episode map shapes during training.}
    \label{fig:diffshape_pinpoint_maze_emptystart_metric_curves_mean_randShape}
\end{subfigure}
\caption{Reward curve of the \convtwo{} model on the \maze{} domain with pinpoints (randomly frozen player and door). On a more challenging task involving randomized per-episode map shapes, the performance gap between models with global and partial observations shrinks.}
\label{fig:diffshape_pinpoint_maze_emptystart_metric_curves_mean}
\end{figure}

To train RL level generators, we use Proximal Policy Optimization (PPO)~\cite{schulman2017proximal} with the same reward function and neural network as in~\cite{khalifa2020pcgrl,earle2021learning}, using the ``narrow'' representation of observations and actions.
In each episode, the model is rewarded by minimizing the loss value between current state and target state (where the user or a training curriculum can vary target solution path length or nearest enemy).
The agent observes an egocentric patch of the level (in which the board may be padded to allow for global observations), and may change the state of its current tile, then is moved to an adjacent tile in an iterative scan of the map.

\subsection{Task}
We extend PCGRL~\cite{khalifa2020pcgrl}, in which level design is framed as a reinforcement learning task. This task is decomposed into a ``problem''---the level design task at hand---and a ``representation''---the interface via which the agent edits the level. 
In this paper, we adapt the \textsl{narrow} representation to support new features, and use the \textsl{binary maze} and \textsl{dungeon} problems and as toy tasks with which to verify our proposed method. 
We initialize the map with the elements of the tasks using the weighted uniform distribution as in~\cite{khalifa2020pcgrl}, or simply leave it empty, depending on the configuration of the environment.
At each timestep, we compute the metrics of interest (i.e. the diameter and number of connected empty regions) if the agent has made any modification to the map.

\paragraph{Binary domain}  
The agent's goal is to create a maze with maximum diameter (i.e. the longest shortest path between any two points in the maze). There are only two types of tile in the maze: wall and air. 
 We approximate the diameter by applying Dijkstra's algorithm twice\footnote{First, we select a random empty tile $x$ in the maze. We apply Dijkstra's algorithm to find the longest shortest path of which $x$ is an endpoint. We then take $y$, the other endpoint of this shortest path, and apply Dijkstra's algorithm starting from $y$ to find the longest shortest path of which $y$ is an endpoint, under the assumption that $y$ is an endpoint of the diameter.} and the number of connected components using a flood fill algorithm.
 
\paragraph{Maze domain}
In this problem, the RL agent needs to use ``wall'' and ``air'' tiles to create a traversable maze from one ``player'' tile to ``door'' tile in the game map. The generated maze should just have one connected component and the solution should be maximized.

\paragraph{Dungeon domain}
Expanding on the \maze{} domain, we consider a task in where the agent's goal is to create a playable dungeon game level. The generated dungeon should have a maximally long shortest path from player, to key, to door. There are 6 types of tile in dungeon problem: wall, air, enemy, key, door, and player. There can be 2 to 5 enemy tiles, and only one key, door, and player tile. Additionally, the distance between the player and the nearest enemy should not be less than 4 tiles. We use Dijkstra's algorithm to find the shortest player-key-door path. 


\begin{table*}
\centering
\adjustbox{max width=\textwidth}{%
\begin{tabular}{lllll|llll}
\toprule
 & \multicolumn{8}{c}{mean ep reward} \\
rand. map shape & \multicolumn{4}{c}{False} & \multicolumn{4}{c}{True} \\
eval map width & 8 & 16 & 24 & 32 & 8 & 16 & 24 & 32 \\
\toprule
obs size &  &  &  &  &  &  &  &  \\
\midrule
15 & 20.43 ± 0.94 & 281.57 ± 74.31 & \textbf{282.40} ± \textbf{1.95} & \textbf{497.95} ± \textbf{6.65} & 6.19 ± 0.33 & 38.48 ± 0.38 & 96.22 ± 2.02 & \textbf{140.16} ± \textbf{4.83} \\
20 & 22.82 ± 0.39 & 193.72 ± 21.75 & 272.26 ± 11.14 & 486.20 ± 20.73 & 7.06 ± 0.62 & 37.56 ± 1.89 & \textbf{97.18} ± \textbf{1.50} & 135.03 ± 2.00 \\
25 & \textbf{24.89} ± \textbf{2.22} & \textbf{338.47} ± \textbf{6.41} & 276.06 ± 13.81 & 474.13 ± 33.43 & \textbf{8.64} ± \textbf{1.70} & \textbf{39.44} ± \textbf{3.18} & 96.55 ± 0.76 & 136.51 ± 1.12 \\
31 & 18.04 ± 0.66 & 269.86 ± 29.98 & 273.20 ± 4.02 & 492.98 ± 6.93 & 7.51 ± 1.08 & 30.35 ± 3.28 & 91.37 ± 2.24 & 132.96 ± 5.04 \\
\bottomrule
\end{tabular}

\label{tab:obss_conv2_dungeon_eval_randomize_map_shap_map_width}
}
\caption{Performance on the \dungeon{} domain, of models with varying observation size. Models are trained on $16\times 16$ maps of fixed shape during training, and evaluated on larger and variable-shaped maps. Smaller observation windows lead to better generalization on these tasks.
}
\label{tab:obss_conv2_dungeon_eval_map_width}
\end{table*}

\begin{figure*}[t]
\centering
\begin{subfigure}{.24\linewidth} 
    \includegraphics[width=\linewidth,trim=0 255 11 0, clip]{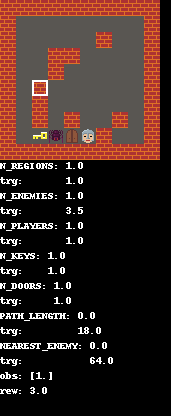}
\end{subfigure}
\begin{subfigure}{.24\linewidth} 
    \includegraphics[width=\linewidth,trim=0 255 0 0, clip]{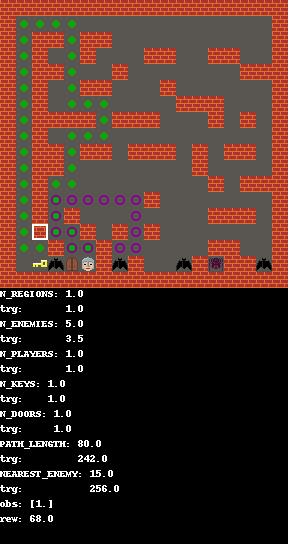}
\end{subfigure}
\begin{subfigure}{.24\linewidth} 
    \includegraphics[width=\linewidth,trim=0 255 0 0, clip]{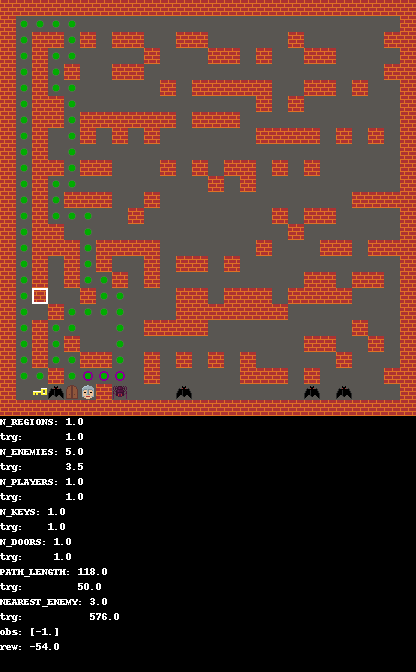}
\end{subfigure}
\begin{subfigure}{.24\linewidth} 
    \includegraphics[width=\linewidth,trim=0 255 0 0, clip]{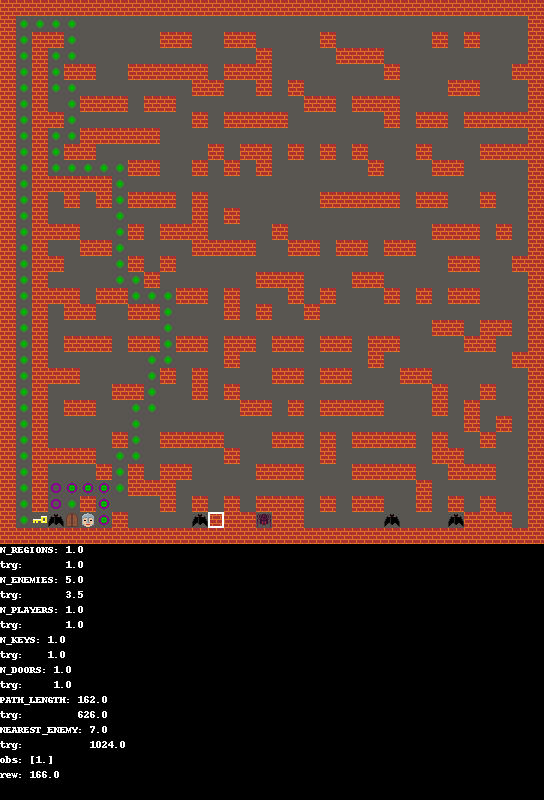}
\end{subfigure}
\caption{On the \dungeon{} domain with controllable path length the \convtwo{} model with $3\times 3$ observation generalizes a design pattern to larger map sizes.}
\end{figure*}

\paragraph{Pinpoint tiles}

We add additional complexity to our tasks, modifying the \maze{} and \dungeon{} domains so that important tiles can be frozen in arbitrary positions on the board. This freezing can be done programmatically during training, or at inference time via a human designer.
These fixed-position tiles provide a more controllable way to prevent agents fromoverfitting to a single optimal output, and allow the problem to become open-looped.


\paragraph{Randomized map shapes}

To train agents that scale to larger and more variable map shapes w.r.t those seen during training, we expose the agent to a variety of map shapes throughout training. So, when training on a $16 \times 16$ map (as in prior PCGRL work) we randomly sample a rectangular map shape from a uniform distribution. The dimensions of this shape are bounded by $3\times 3$ and $16 \times 16$. Another alternative might be to progressively train on maps of increasing size. Although intuitive, this approach introces two failrue modes. First, it risks catastrophic forgetting of smaller map sizes seen earlier during training. Conversely, it may incentivise the agent to learn faulty representations on smaller maps that do not transfer to larger ones.



\subsection{Jax implementation}

We use the jax python library~\cite{jax2018github} to implement our PCGRL environments and training algorithm. Jax exposes a wide variety of tensor-based operations to the user, mirroring much of the functionality of numpy and/or libraries like pytorch or tensorflow, compiling operations to XLA just-in-time. Provided that the size of tensors is fixed at compile-time, and with some limitations on logical operations like \texttt{if} conditions, Jax can improve runtime efficiency by ``fusing'' lower-level operations.
Our Jax implementation of PCGRL builds on gymnax~\cite{lange2022gymnax} and the PureJaxRL code base~\cite{purejaxrl}, which implements a number of simple embodied game-playing environments in Jax.


To pathfind in jax, we can flood activation out to adjacent traversible tiles in parallel across the board using convolutional kernels. 
Similar logic can be used to compute the number of regions.




\section{Results}

\paragraph{Speed Comparison}

In \autoref{tab:jax_train_fps} we calculate the FPS during training of the prior CPU implementation (splitting environments across 11 cores), and jax-pcgrl respectively.
We find that the jax implementation leads to speedups of over $15\times$ relative to the CPU implementation.
While the CPU implementation plateaus as the number of environments reaches between 50-100, the FPS of the jax version continues to increase significantly up to at least $600$ workers.

In \autoref{tab:jax_fps}, we calculate the speed of our jax implementations of the \binary{} \maze{} and \dungeon{} domains in frames per second given actions from a random agent, achieving 45k FPS.

\paragraph{Observation size}
In the following experiments, we train the agents in a fixed size map setting and evaluate them on different map shapes. From these thorough experiment configurations, we can answer the question about how observation size influence the model performance on in-distribution and out-distribution map and the how well the agents generalize.

In \autoref{tab:obss_conv2_dungeon_eval_map_width} we evaluate the performance of the \convtwo{} model on the \dungeon{} task. 
We evaluate different obseravtion input of the model on varying maximum random map widths and random per-episode map shapes.
Performance of models with smaller observation windows is slightly weaker than models with full observations on in-distribution maximum map widths.
But smaller observations lead to better generalization to larger maximum map widths, despite these models having fewer parameters then their fully-observing counterparts.

\begin{table*}
\centering
\adjustbox{max width=\textwidth}{%
\begin{tabular}{lllll|llll}
\toprule
 & \multicolumn{8}{c}{mean ep reward} \\
rand. map shape & \multicolumn{4}{c}{False} & \multicolumn{4}{c}{True} \\
eval map width & 8 & 16 & 24 & 32 & 8 & 16 & 24 & 32 \\
\toprule
obs size &  &  &  &  &  &  &  &  \\
\midrule
3 & \textbf{23.77} ± \textbf{2.21} & 175.03 ± 14.62 & \textbf{318.68} ± \textbf{24.37} & \textbf{548.01} ± \textbf{36.50} & 7.70 ± 1.56 & 35.65 ± 2.71 & 99.90 ± 4.11 & 138.22 ± 3.69 \\
5 & 22.66 ± 1.24 & 175.48 ± 2.56 & 288.45 ± 0.68& 514.73 ± 2.17 & 8.40 ± 1.46 & \textbf{37.56} ± \textbf{1.74} & \textbf{100.57} ± \textbf{5.48} & 139.29 ± 3.46 \\
8 & 19.69 ± 0.71 & 140.76 ± 2.66 &285.70 ± 0.67& 509.28 ± 2.80 & 7.23 ± 0.88 & 36.82 ± 1.16 & 99.73 ± 1.72 & \textbf{145.10} ± \textbf{4.09} \\
16 & 18.53 ± 2.14 & 183.15 ± 13.76 & 282.80 ± 4.51& 490.29 ± 6.66 & 5.34 ± 0.88 & 36.46 ± 1.59 & 98.90 ± 3.11 & 136.70 ± 1.52 \\
31 & 20.40 ± 0.55 & \textbf{184.12} ± \textbf{28.38} & 273.73 ± 0.69 & 492.11 ± 1.62 & \textbf{8.60} ± \textbf{0.85} & 35.48 ± 0.38 & 93.54 ± 2.61 & 136.60 ± 3.06 \\
\bottomrule
\end{tabular}

\label{tab:obss_dungeon_conv2_ctrl_path_eval_randomize_map_shap_map_width}
}
\caption{Performance on the \dungeon{} domain, with controllable path length, of the \convtwo{} model with varying observation size. Smaller observation windows often lead to better generalization.}
\label{tab:obss_dungeon_conv2_ctrl_path_eval_map_width}
\end{table*}

\begin{table*}
\centering
\adjustbox{max width=\textwidth}{%
\begin{tabular}{lllll|llll}
\toprule
 & \multicolumn{8}{c}{mean ep reward} \\
rand. map shape & \multicolumn{4}{c}{False} & \multicolumn{4}{c}{True} \\
eval map width & 8 & 16 & 24 & 32 & 8 & 16 & 24 & 32 \\
\toprule
obs size hid dims &  &  &  &  &  &  &  &  \\
\midrule
3 & \textbf{37.21} ± \textbf{9.46} & 188.29 ± 8.83 & \textbf{372.66} ± \textbf{3.49} & \textbf{582.95} ± \textbf{30.90} & \textbf{10.89} ± \textbf{2.20} & 39.89 ± 1.92 & \textbf{103.75} ± \textbf{0.95} & 141.59 ± 2.10 \\
5 & 24.26 ± 0.89 & 193.17 ± 17.19 & 295.83 ± 6.47 & 517.08 ± 0.72 & 9.87 ± 0.60 & 41.39 ± 3.70 & 103.48 ± 5.41 & \textbf{145.19} ± \textbf{5.50} \\
8 & 27.00 ± 0.78 & 181.41 ± 15.48 & 316.44 ± 17.98 & 520.65 ± 6.83 & 8.53 ± 0.17 & 38.99 ± 2.88 & 103.19 ± 1.98 & 139.42 ± 0.81 \\
16 & 19.25 ± 0.72 & 182.88 ± 8.26 & 265.95 ± 11.06 & 451.51 ± 24.09 & 5.80 ± 0.14 & \textbf{44.19} ± \textbf{3.45} & 100.01 ± 6.53 & 135.75 ± 6.24 \\
31 & 17.43 ± 0.77 & \textbf{206.98} ± \textbf{21.27} & 276.81 ± 5.06 & 490.94 ± 15.53 & 5.03 ± 0.81 & 31.03 ± 1.30 & 92.21 ± 3.50 & 133.53 ± 1.93 \\
\bottomrule
\end{tabular}

\label{tab:obss_hiddims_dungeon_conv2_ctrl_path_eval_randomize_map_shap_map_width}
}
\caption{Performance on the \dungeon{} domain, with controllable path length, of the \convtwo{} model with varying observation size---while fixing the number of learnable model parameters. Relative to \autoref{tab:obss_dungeon_conv2_ctrl_path_eval_map_width}, increasing the number of parameters in partially-observing models leads to improvements on certain out-of-distribution scenarios (though not on randomized map shapes).}
\label{tab:obss_hiddims_dungeon_conv2_ctrl_path_eval_map_width}
\end{table*}

In \autoref{tab:obss_dungeon_conv2_ctrl_path_eval_map_width}, we evaluate the \convtwo{} model on the \dungeon{} domain with random target path length for learning controllability.
Looking at evaluation scenarios without randomized per-episode map shapes, we see that on in-distribution $16\times 16$ maps, global observations perform the best. On larger maps, however, more local partial observations generalize better, despite the fact that models with smaller observation windows have significantly fewer parameters.
When evaluating with randomized per-episode map shapes, however, models of all observation sizes generalize comparably.

\begin{figure*}
\begin{subfigure}[t]{.5\linewidth}
    \includegraphics[width=\linewidth]{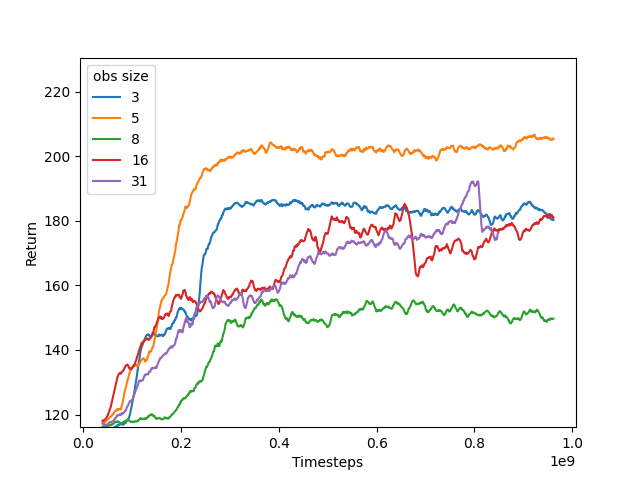}
    \caption{Models with smaller observation windows have fewer parameters.}
    \label{fig:obss_dungeon_conv2_ctrl_path_metric_curves_mean_noHidDims}
\end{subfigure}
\begin{subfigure}[t]{.5\linewidth}
    \includegraphics[width=\linewidth]{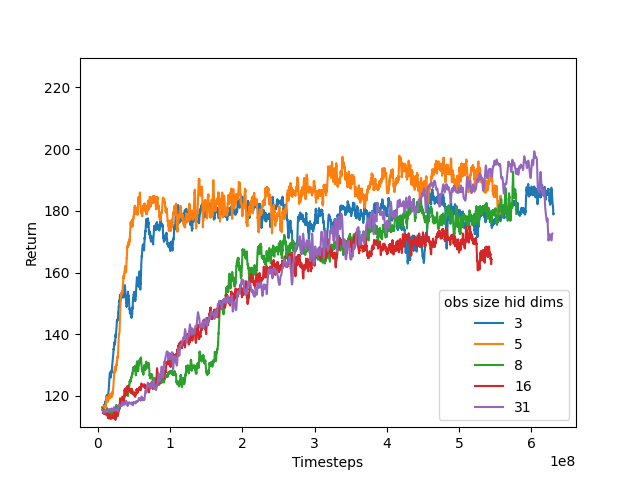}
    \caption{Models with smaller observation windows have greater hidden dimensions and a similar number of learnable parameters.}
    \label{fig:obss_hiddims_dungeon_conv2_ctrl_path_metric_curves_mean_hidDims}
\end{subfigure}
\caption{Effect of different observation sizes on reward curves during training of the \convtwo{} model on the \dungeon{} domain with control targets.}
\label{fig:obss_hiddims_dungeon_conv2_ctrl_path_metric_curves_mean}
\end{figure*}

We repeat this experiment in \autoref{tab:obss_hiddims_dungeon_conv2_ctrl_path_eval_map_width}, where we systematically add hidden nodes to each layer of each network until it has almost as many (but no more) learnable parameters than the model with full observations---effectively separating the effect of partial observations from model size.
Here, we see that models with smaller observation size but comparable numbers of parameters tend to outperform models with larger observations across all maximum map sizes.

In \autoref{fig:obss_hiddims_dungeon_conv2_ctrl_path_metric_curves_mean}, we plot the reward curves of the experiments in \autoref{tab:obss_dungeon_conv2_ctrl_path_eval_map_width} and \autoref{tab:obss_hiddims_dungeon_conv2_ctrl_path_eval_map_width} during training, and similarly observe that the addition of learnable parameters to models with smaller observation sizes improves their performance. \todo[]{I still don't understand this figure lol why is purple better on the right?!}

\begin{table*}
\centering
\adjustbox{max width=\textwidth}{%
\begin{tabular}{llllll|llll}
\toprule
 &  & \multicolumn{8}{c}{mean ep reward} \\
 & rand. map shape & \multicolumn{4}{c}{False} & \multicolumn{4}{c}{True} \\
 & eval map width & 8 & 16 & 24 & 32 & 8 & 16 & 24 & 32 \\
\toprule
rand. map shape & obs size &  &  &  &  &  &  &  &  \\
\midrule
\multirow[t]{4}{*}{False} & 8 & \textbf{7.18} ± \textbf{0.97} & 22.45 ± 6.02 & \textbf{44.57} ± \textbf{4.79} & \textbf{59.07} ± \textbf{3.78} & \textbf{1.23} ± \textbf{0.59} & \textbf{4.82} ±\textbf{ 1.59} & \textbf{9.57} ± \textbf{4.40} & \textbf{6.10 }± \textbf{7.43} \\
 & 16 & 3.93 ± 4.43 & 48.17 ± 2.70 & 39.53 ± 9.98 & 20.08 ± 8.51 & 0.27 ± 0.48 & -1.28 ± 4.50 & -3.15 ± 2.87 & -8.14 ± 3.09 \\
 & 24 & 2.31 ± 1.42 & 41.87 ± 1.70 & 18.55 ± 7.21 & 0.43 ± 8.78 & 0.87 ± 0.64 & 1.07 ± 1.08 & 1.11 ± 0.92 & 0.79 ± 0.46 \\
 & 31 & 3.19 ± 0.95 & \textbf{51.37} ± \textbf{7.41} & -10.18 ± 5.56 & -14.73 ± 9.70 & 0.93 ± 0.03 & -2.33 ± 3.87 & -5.97 ± 5.22 & -6.71 ± 6.92 \\
\cline{1-10}
\multirow[t]{4}{*}{True} & 8 & \textbf{9.24} ± \textbf{0.30} &  \textbf{22.13} ±  \textbf{3.08}& \textbf{29.88} ± \textbf{9.93} & \textbf{30.48} ± \textbf{9.47} & \textbf{6.61} ± \textbf{0.16} & 20.09 ± 2.70 & \textbf{26.11} ± \textbf{4.08} & \textbf{25.12} ± \textbf{1.85} \\
 & 16 & 5.36 ± 1.36 & 13.38 ± 2.82 & 0.42 ± 4.86 & -9.63 ± 6.02 & 5.93 ± 0.39 & \textbf{21.75} ± \textbf{1.76} & 16.57 ± 2.79 & 16.87 ± 4.97 \\
 & 24 & 5.05 ± 1.13 & -3.97 ± 4.97 & -19.45 ± 5.74 & -31.21 ± 16.32 & 5.08 ± 0.76 & 18.31 ± 1.93 & 5.67 ± 3.70 & -0.84 ± 3.65 \\
 & 31 & 4.25 ± 1.43 & -1.48 ± 1.82 & -9.69 ± 7.61 & -16.08 ± 11.71 & 3.76 ± 0.21 & 18.69 ± 0.55 & 3.82 ± 1.73 & -0.68 ± 2.19 \\
\cline{1-10}
\bottomrule
\end{tabular}
}

\label{tab:diffshape_pinpoint_maze_emptystart_eval_randomize_map_shap_map_width}
\caption{Performance on the pinpointed \maze{} domain, of the \convtwo{} model, with varying observation size and with or without exposure to randomized map shapes during training. When map shapes are randomized during training, the in-distribution performance gap between local and global models 
closes.}
\label{tab:diffshape_pinpoint_maze_emptystart_eval_randomize_map_shap_map_width}
\end{table*}

\paragraph{Randomized map shapes during training}
In \autoref{tab:diffshape_pinpoint_maze_emptystart_eval_randomize_map_shap_map_width}, we examine the performance of the \convtwo{} model on the \maze{} domain, with pinpoints (randomized fix placement of player and door tiles).
In these experiments, we show how randomizing the map shape during training will affect the model performance of generalization under different observation size.

First, we look at evaluation of models on fixed square map sizes.
In this setting, models exposed to similarly fixed square map shapes during training tend to outperform those trained on variable per-episode map shapes.
Next, we evaluate models while randomly sampling shapes within these maximum sizes on each evaluation episode (right side of the table).
Models trained without randomized map shapes are broadly unable to adapt to variable map shapes during evaluation---with the only consistent positive reward in this setting coming from models with the smallest observation window ($8\times 8$).
Of the models trained on variable map shapes, smaller observation windows generally outperform larger ones.

In \autoref{fig:diffshape_pinpoint_maze_emptystart_metric_curves_mean}, we plot reward curves of these models.
When map shapes are fixed between episodes (\autoref{fig:diffshape_pinpoint_maze_emptystart_metric_curves_mean_fixedShape}), models with full observations outperform those with local observations.
But when map shapes are randomized over each episode (\autoref{fig:diffshape_pinpoint_maze_emptystart_metric_curves_mean_randShape}), smaller observation windows lead to comparable or better performance than those with full observations.
This trend is also replicated at evaluation time in \autoref{tab:diffshape_pinpoint_maze_emptystart_eval_randomize_map_shap_map_width}, and can be observed qualitatively in \autoref{diffshape_pinpoint_maze_emptystart_maps}

\begin{figure}[t]
\begin{subfigure}{\linewidth} 
    \includegraphics[width=\linewidth,trim=0 705 20 0, clip]{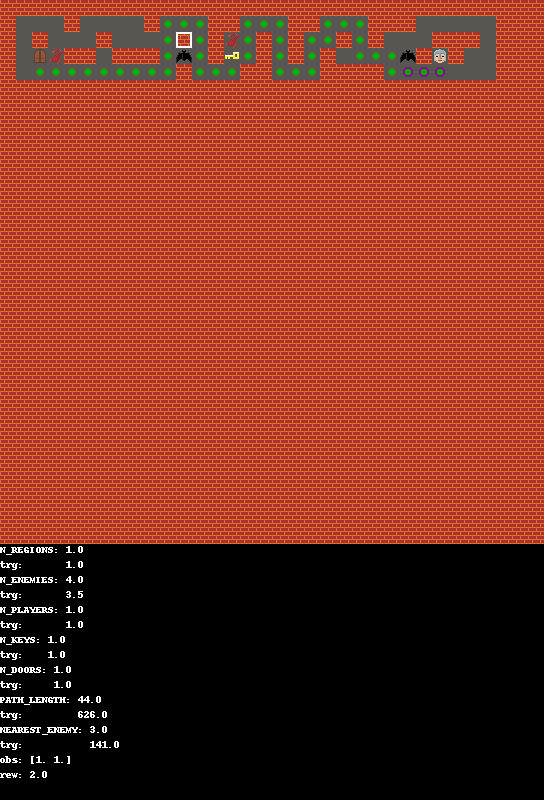}
\end{subfigure}
\begin{subfigure}{.38\linewidth}
\vspace{7pt}
    \includegraphics[width=\linewidth,trim=0 320 50 0, clip]{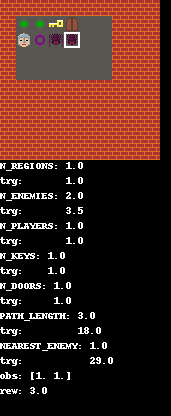}
\end{subfigure}
\begin{subfigure}{.6\linewidth} 
\vspace{7pt}
    \includegraphics[width=\linewidth,trim=0 570 220 0, clip]{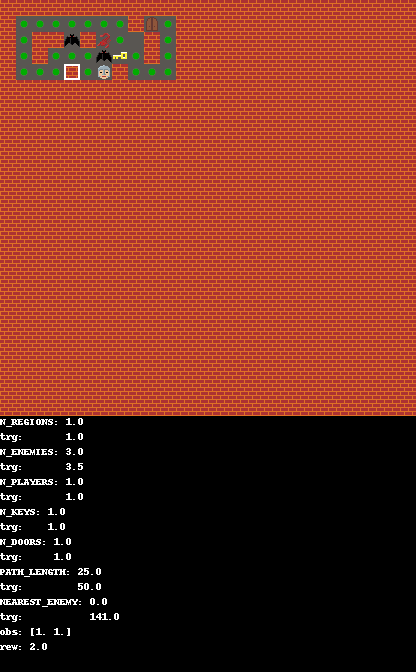}
\end{subfigure}
\begin{subfigure}[t]{.28\linewidth} 
\vspace{1pt}
    \includegraphics[width=\linewidth,trim=0 255 130 0, clip]{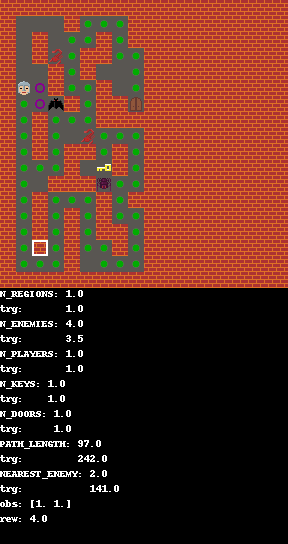}
\end{subfigure}
\begin{subfigure}[t]{.68\linewidth} 
\vspace{1pt}
    \includegraphics[width=\linewidth,trim=0 255 0 0, clip]{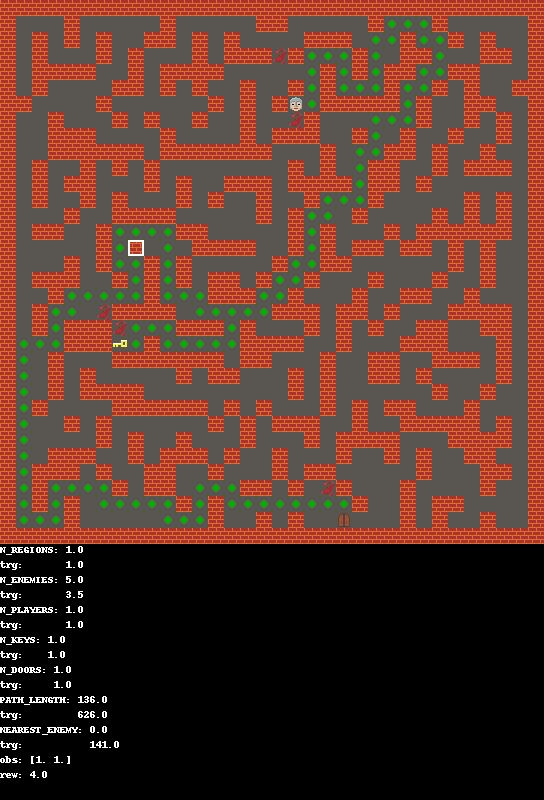}
\end{subfigure}
\caption{A model taking local $8\times 8$ observations generalizes a design strategy across varied map shapes.}
\end{figure}

\section{Discussion}

To the best of our knowledge, no prior jax RL environments involving path-finding, which could also be a useful addition to player environments (e.g. to simulate enemy navigation in \cite{matthews2024craftax}).
If the speedups~\autoref{tab:jax_fps} of our jax implementations of PCGRL environments are less than those of other, still simpler environments, this might come from the added complexity of our path-finding implementation.

\autoref{tab:diffshape_pinpoint_maze_emptystart_eval_randomize_map_shap_map_width} shows us that in general, smaller observation windows lead to higher performance on out-of-distribution settings, either for larger scale or per-episode map shapes.
We contend that this is likely a result of overfitting under global observation: models that are accustomed to seeing the padding of unique ``border'' tiles surrounding the effective map region are disrupted when, during evaluation on larger maps, these tiles are suddenly not present in their egocentric observations.
These models are likely using the placement of these border tiles to infer global coordinates, allowing it to consistently construct one optimal global level (or a set of global levels, when dealing with randomized map shapes, controllable path-length metrics, or frozen ``pinpoint'' tiles).
With such global coordinates at hand, these levels can theoretically be generated in one shot (or one scan over the board).

By restricting the observation space, on the other hand, models could only infer global coordinates by editing the entire board multiple times, communicating relative coordinates via patterns that cascade across the map in an iterative way.
The fact that these local-observation models perform better out of distribution suggests that such an approach to iteratively passing local information across the board leads to more general representations and strategies for designing good levels.
In other words, these constrained models are less likely to memorize \textit{what} one or a set of optimal levels should look like, and instead may learn general strategies for \textit{how} to improve or modify maps along certain axes.

\todo[inline]{Refer to figures of level generation to support this argument.}

Meanwhile, models trained on fixed-size square maps are not able to adapt well to per-episode variation of map shape.
Models with full observations exhibit particularly pronounced failure in this case, and our reasoning would expect that a model that has observed only square border shapes is disrupted when it observes rectangular shapes during evaluation.
But models with local observations are also thrown off by variable per-episode map shapes, suggesting that their strategies for iteratively transmitting local information are not robust to non-square map shapes.
Conversely, maps trained on variable per-episode map shapes do not quite attain the performance of models trained strictly on fixed-size square maps---even on out-of-distribution sizes.
We expect that this is merely a result of insufficient training time given the larger task distribution on which these models are trained, and that further training would allow them to better cover this distribution with good performance.

The reward curves in \autoref{fig:diffshape_pinpoint_maze_emptystart_metric_curves_mean}, along with the in-distribution columns of \autoref{tab:diffshape_pinpoint_maze_emptystart_eval_randomize_map_shap_map_width}, reveal that when per-episode map shapes are randomized, models taking full observations seem to lose the advantage they have when map shapes are fixed.
This would seem to suggest that knowing the overall shape of the map is actually a disadvantage, even on in-distribution tasks, when this distribution is diverse enough.
In other words, we hypothesize that models that are forced to learn general level-editing strategies, adaptive to a range of possible map shapes, arrive more quickly at optimal performance on the set of training map shapes, while models that have access to this information are effectively distract, and drawn away from more general and robust strategies.
Or, when the training distribution is wide enough, limiting a model's direct access to information about precisely which training task it is in at a given moment can render it more effective, because this model is forced to find similarities between tasks and effectively learn compressed representations of this distribution.




\section{Conclusion}

The over $15\times$ speedups achieved by our jax reimplementions of PCGRL environments allow us to train models for many more timesteps (1 billion) than in previous works (around 200 million).
By randomizing map shapes and the placement of pivotal items in initial layouts, we allow for training of more robust and controllable level generator agents.
In our experiments, we find that limiting the observation window of trained agents leads to stronger generalization.
By evaluating on held-out initial map layouts and constraints, pcgrl-jax can serve as a scalable benchmark for RL agents, with real-world applications for human level designers.

\bibliography{ref}

\appendix

\begin{figure}[t]
\centering
\begin{subfigure}{.24\linewidth} 
    \includegraphics[width=\linewidth,trim=0 105 140 0, clip]{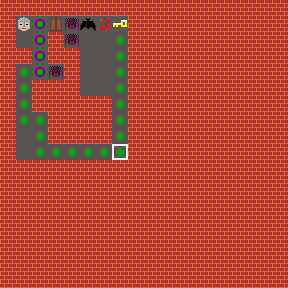}
\end{subfigure}
\begin{subfigure}{.24\linewidth} 
    \includegraphics[width=\linewidth,trim=0 105 140 0, clip]{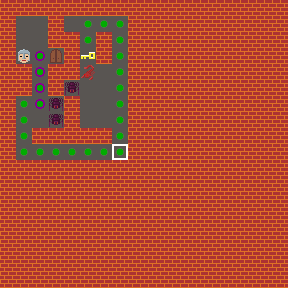}
\end{subfigure}
\begin{subfigure}{.24\linewidth} 
\vspace{4pt}
    \includegraphics[width=\linewidth,trim=0 105 140 0, clip]{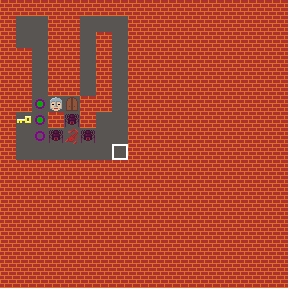}
\end{subfigure}
\begin{subfigure}{.24\linewidth} 
    \includegraphics[width=\linewidth,trim=0 105 140 0, clip]{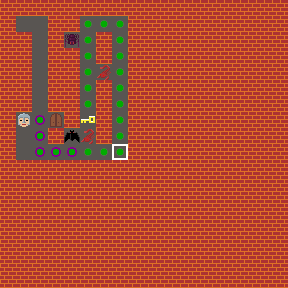}
\end{subfigure}
\caption{On the \dungeon{} domain the \convtwo{} model with full observation mutates the board through a series of playable levels.}
\end{figure}

\end{document}